# Classification of Seeds using Domain Randomization on Self-Supervised Learning Frameworks

Venkat Margapuri and Mitchell Neilsen

*Abstract*— The first step toward Seed Phenotyping i.e. the comprehensive assessment of complex seed traits such as growth, development, tolerance, resistance, ecology, yield, and the measurement of parameters that form more complex traits is the identification of seed type. Generally, a plant researcher inspects the visual attributes of a seed such as size, shape, area, color and texture to identify the seed type, a process that is tedious and labor intensive. Advances in the areas of computer vision and deep learning have led to the development of convolutional neural networks (CNN) that aid in classification using images. While they efficiently perform classification, a key bottleneck is the need for an extensive amount of labelled data to train the CNN before it can be put to the task of classification. The accumulation of a plethora of data is a challenge, let alone labeling it. The work strives to address the aforementioned challenges and leverages the concepts of Contrastive Learning and Domain Randomization in order to achieve the same. Briefly, domain randomization is the technique of applying models trained on images containing simulated objects to real-world objects. The use of synthetic images generated from a representational sample crop of real-world images alleviates the need for a large volume of test subjects, seeds in this case. Contrastive learning is a self-supervised learning paradigm that works by identifying distinctiveness among test subjects. The idea is that the encodings of similar objects are similar and dissimilar objects, dissimilar. As part of the work, synthetic image datasets of five different types of seed images namely, canola, rough rice, sorghum, soy and wheat are applied to three different self-supervised learning frameworks namely, SimCLR, Momentum Contrast (MoCo) and Build Your Own Latent (BYOL) where ResNet-50 is used as the backbone in each of the networks. The performance of the self-supervised learning frameworks is compared against the performance of a supervised learning model built on ResNet-50. When the self-supervised models are fine-tuned with only 5% of the labels from the synthetic dataset, results show that MoCo, the model that yields the best performance of the self-supervised learning frameworks in question, achieves an accuracy of 77% on the test dataset which is only ~13% less than the accuracy of 90% achieved by ResNet-50 trained on 100% of the labels. The experiment demonstrates the feasibility of domain randomization for classification of seeds using self-supervised learning frameworks.

*Keywords*—BYOL, Contrastive Learning, Domain Randomization, Momentum Contrast (MoCo), ResNet-50, SimCLR

## I. Introduction

Agriculture is one of the leading sectors of economy in the world. Agriculture, food, and related industries contributed $1.109 trillion to the U.S. GDP in 2019, a 5.2 percent share [34]. The use of agriculture in technology started over a century ago, and, to improve production efficiency, several studies have been conducted since the 1990s [29]. One of the key areas of study is High Throughput Phenotyping (HTP) in seeds, commonly known as Seed Phenotyping. Seed Phenotyping is the comprehensive assessment of complex seed traits such as growth, development, tolerance, resistance, ecology, yield, and the measurement of parameters that form more complex traits [15]. The primary step in the process of seed phenotyping is the identification of the seed type i.e. classification of seeds. Seed type is generally identified by trained plant scientists in a laboratory setting. However, the availability of a plethora of knowledge and resources in the seed domain led to a spike in the number of citizen scientists working toward seed phenotyping. In order to assist such citizen scientists further their research and also to help validate the results of trained plant scientists, automated systems that aid in the classification of seeds are essential. Convolutional neural networks (CNN) are one such solution to aid in classification. The process of building classifiers with CNNs generally involves training the CNN on a large volume of labeled data where the subjects are real-world entities. However, the procurement of such labeled data can be challenging, especially for citizen scientists who don't always have the luxury of a laboratory setting. The work focuses to address two aspects – a) the need for a large volume of real-world entities to train neural networks b) the need for labeled data to train neural networks. In order to mitigate the need for a large volume of real-world entities, the work proposes the application of domain randomization i.e. the technique of training models on simulated images that transfer to real images. Contrastive self-supervised learning frameworks are suggested to help train the CNNs without an abundance of labeled data. The work evaluates the performance of three self-supervised learning frameworks namely, SimCLR, Momentum Contrast (MoCo) and BYOL against the performance of a supervised learning model using ResNet-50 on synthetic image datasets. The synthetic image datasets are generated using a randomly chosen seed sample of each of the seeds of canola, rough rice, sorghum, soy and wheat.

In the remainder of the paper, section II discusses the related work, section III describes the intuition behind contrastive self-supervised learning, section IV discusses the experiment, section V, results and section VI concludes the article.

## II. Related Work

Relevant research involving a combination of domain

Venkat Margapuri is with Kansas State University (corresponding author, e-mail: marven@ksu.edu).

Mitchell Neilsen is with the Department of Computing and Information Sciences, Kansas State University, Manhattan, KS - 66502 USA (e-mail: neilsen@ksu.edu).

randomization and self-supervised learning is not abundant. However, works that employ domain randomization and self-supervised learning independently are available. In terms of training neural network models using synthetic datasets, Toda et. al [31] applied synthetic datasets to Mask RCNN, an instance detection framework, to perform instance segmentation and detection of seeds. Gulzar et. al [9] developed a seed classification system for 14 different types of seeds by applying transfer learning on VGG-16, a convolutional neural network.

In the realm of contrastive self-supervised learning frameworks, Chen et. al [6] proposed the SimCLR framework, a major breakthrough in terms of contrastive self-supervised learning frameworks in the sense that don't require a memory bank. The framework relies on the presence of large batches of training data. The work demonstrated the critical role played by the composition of data augmentations on contrastive learning. He et. al [10] proposed the Momentum Contrast (MoCo) framework for unsupervised visual representation learning that approached contrastive learning as a dictionary look-up problem. A dynamic dictionary with a queue and a moving-averaged encoder is built on-the-fly to facilitate contrastive unsupervised learning. Chen et. al [5] proposed a modified framework to MoCo by using a multi-layer perceptron projection head and more data augmentation that outperformed SimCLR without the need for large training batches. Grill et. al [8] introduced the Bootstrap Your Own Latent (BYOL) approach to self-supervised image representation learning. BYOL works by the use of an online and a target neural network. The principle is to train the online network on an augmented view of an image to predict the target network representation of the image that is a different augmented view.

### III. SELF-SUPERVISED LEARNING

Self-Supervised learning refers to the technique of training neural networks where the training data is labeled automatically [4]. Self- supervised learning consists of the traits of both supervised and unsupervised learnings. It is supervised in the sense that the model is still trained to learn a function from pairs of inputs and labeled outputs. It is unsupervised in the sense that the model learns without being provided with labels. Contrastive Self-Supervised learning falls under the broader umbrella of self-supervised learning where the intuition is to learn the features of a dataset by a measure of distinctiveness. While labels are absent, the model learns the similarity between the datapoints in the dataset and groups them together. This process results in obtaining groups of similar datapoints, yielding in classification.

Three frameworks namely, SimCLR, Momentum Contrast (MoCo) and Build Your Own Latent (BYOL) are used as part of the experiments. The architecture and working of each of the frameworks is as described further.

#### A. SimCLR

SimCLR functions by maximizing the similarity measure between two augmented views obtained from the same image. The framework is explained as a three-step process:

1. **Data Augmentation:** Given an input image, two correlated views of the image are generated using three different transformations namely, random crop and resize, horizontal flip and color distortion involving jitter and grayscale conversion. This process helps in generating different views of the same image aiding the network in its ability to be transformation independent.
2. **Neural Network Encoder:** The augmented images are put through an encoder such as ResNet-18 or ResNet-50 and vector encodings of the images are extracted.
3. **Projection Head:** A neural network encoder such as ResNet-50 generates encodings that are 2048 dimensional. In order to reduce the dimensionality of the vector encodings, a multi-layer perceptron (MLP) with one hidden layer is used. The encodings in high dimensional space are mapped to 128-dimensional latent space upon which contrastive loss is applied. The activation function used in the MLP is ReLU. The architecture of SimCLR is as shown in Fig. 1.

**Contrastive Loss:** Given a set of augmented images that contains two correlated views for each of the images in the set, the contrastive loss functions aims to identify each pair of correlated views that belong to the same image. The vector encodings in 128-dimensional latent space are run through the contrastive loss function that takes similarity of the images into account. The similarity between two images is computed using a measure named Cosine Similarity. It is defined as $sim(x,y) = (x^t y)/|x||y|$ where x and y are two different vector encodings. The contrastive loss function that takes cosine similarity into account is defined as:

$$L_{i,j} = -log \frac{e^{\frac{sim(z_i, z_j)}{\tau}}}{\sum_{k=1}^{2N} e^{\frac{sim(z_i, z_k)}{\tau}}_{k \neq i}}$$

$Z = \{z_1, z_2, …, z_k\} \in \mathbb{R}^k$ are output vectors from the projection head. In the event that two vectors are similar, the function yields a result of zero which is the optimal loss. $\tau$ is the temperature parameter used to scale the cosine similarities. Reference [6] found that the optimal temperature parameter helps the model learn from hard negatives. It is worth noting that the value of $\tau$ depends on the number of epochs and batch size during training.

Given a batch of N images, the augmentation results in each image having two representations yielding a total of 2N augmented images. The positive pair is the pair of images that are a result of augmentation from the same image and every other pair in the set is considered a negative pair. The augmented image set consists of one positive pair and $2(N – 1)$ negative pairs. The framework relies on the presence of large models and batch sizes to achieve better accuracy. This requirement is also deemed a bottleneck for the framework because the increase in batch and model sizes have a direct impact on the amount of computational resources required to train the network.

#### B. MoCo

Momentum Contrast (MoCo) is a technique similar to SimCLR with the key improvement being that it eliminates the need for large models and batch sizes employed by SimCLR. There is one key difference that sets SimCLR and MoCo apart from each other; the use of two neural network encoders. MoCo functions on the principle of matching queries to keys where key and query refer to the encodings of an augmented image. As opposed to the idea of having a single neural network encoder in SimCLR, a second neural network encoder, similar to the first is introduced where one of them generates the encodings for the key and the other, query. MoCo constructs a dictionary built and operated as a queue that keeps a history of the encoded keys. The dictionary in turn, acts as the resource pool for positive and negative pairs of images. A positive pair refers to the instance where a query matches the key. Every other sample in the dictionary that doesn't correspond to the query acts as a negative sample.

One of the key challenges of the approach is that, learning the parameters of the key encoder requires calculating the gradients of each of the samples in the queue. The larger the number of samples, the greater the amount of computational resources required. In order to address the issue, MoCo updates the key encoder with the momentum-based average of the query encoder. It is defined as $\theta_k \leftarrow m\theta_k + (1-m)\theta_q$ where $\theta_k$ and $\theta_q$ are the parameters of key and query encoders respectively, m is the momentum whose value is kept close to one. The architecture of MoCo is as shown in Fig. 2.

### C. Bootstrap Your Own Latent (BYOL)

BYOL is a technique similar to SimCLR and MoCo with one key difference. BYOL does not take negative samples into account. Instead, BYOL focuses solely on ensuring that similar samples have similar representations. While it might not be apparent at first glance as to the reason it is significant, it is important to understand that BYOL avoids the collapsed representation problem without negative samples. Collapsed representation is the state wherein a network trained only on similar pairs learns a constant function since the loss output over similar pairs is always a constant, such as zero. No discriminative features are learnt rendering the network unfit for prediction on a test dataset or fine-tuning on a different dataset. The working of BYOL is described as follows:

1. **Target and Online Networks:** Consider two encoders with the same architecture, generally ResNet-50, where in one of them is randomly initialized to be the 'target' network and another set to be trainable known as the 'online' network.
2. **Data Augmentation:** Pass an input image 'I' through the data augmentation pipeline to generate two stochastically augmented views 'I1' and 'I2'.
3. **Vector Encoding Extraction:** Run the augmented views 'I1' and 'I2' through the 'target' and 'online' networks respectively and extract the vector representations of the views
4. **Projection Head:** Use an MLP to reduce the vector encodings to a lower dimensional latent space, usually 256.
5. **Prediction:** The vector encoding from the 'online' network is used to predict the vector encoding from the 'target' network i.e. the distance between the vector encodings is minimized using the normalized mean squared error loss function.
6. **Update Target Network:** The 'target' network is updated at the end of each training step as the exponential moving average of the parameters of the 'online' network. The 'target' network closely follows yet lags behind the 'online' network at all times.

In essence, BYOL attempts to leverage the 'online' network in one training step as the 'target' network in the subsequent training step. Doing so avoids the collapsed representation problem and also betters the performance of the network. The architecture is as shown in Fig. 3.

## IV. DATASET AND METHOD

### A. Dataset

The seeds of canola, rough rice, sorghum, soy and wheat are used as part of the experiment. Since the work attempts to study the feasibility of domain randomization on self-supervised learning frameworks, the images used for the training and validation datasets are synthetic images generated using a representative sample of each of the seeds. A large seed pool of the order of several hundred is available considering the nature of research conducted in the laboratory is primarily focused on agriculture. 30 seeds belonging to each of the seed types in consideration are chosen at random and photographed individually resulting in a sample of 150 images. In accordance with the technique outlined in a prior work [18], images containing seeds overlaid on a common background are generated where in each image contains 50 seeds of a certain seed type as shown in Fig. 4. 1000 images per seed type, each 224 x 224 x 3 in size, are generated of which 800 are used for training and 200 for validation. As for the test dataset, 40 images of each seed type are photographed where in each of the images contains ~50 seeds of the same seed type placed on a lightbox. It is ensured that the seeds in the test dataset are not the seeds used to compile the synthetic image dataset. Overall, the training dataset comprises of 5000 images while the test dataset is made up of 200 images.

### B. Experimental Setup

ResNet-50 is used as the encoder for each of the self-supervised learning frameworks in consideration, SimCLR, MoCo and BYOL. The implementation of the frameworks is done in PyTorch and executed on a Tesla V100 GPU provided by Google Colab Pro. The lr-finder module is used to find the learning rate that best fits a model for fine-tuning.

### C. SimCLR, MoCo and BYOL

Each of the frameworks, SimCLR, MoCo and BYOL, is implemented as described in section III using a ResNet-50 model pre-trained on the ImageNet dataset as encoder. Each of the frameworks is trained for a total of 50 epochs at a learning rate of $1e^{-3}$ on the training dataset. Weight Decay, a regularization technique, set to $1e^{-4}$ is added to the models to prevent overfitting. It is worth noting that the training at this stage does not involve any labels and is solely to minimize the loss functions of each of the frameworks. The implementation details specific to each of the frameworks are described further.

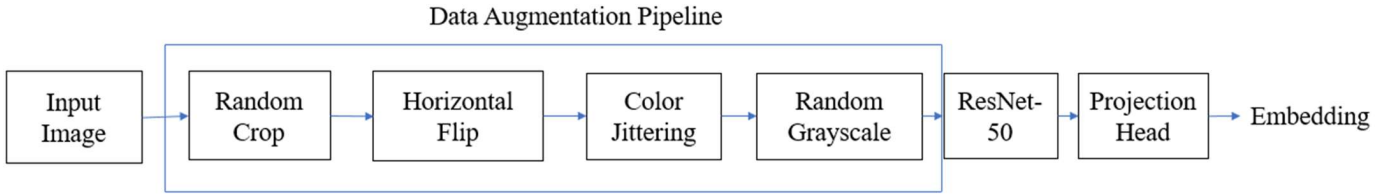
Fig. 1: SimCLR Architecture

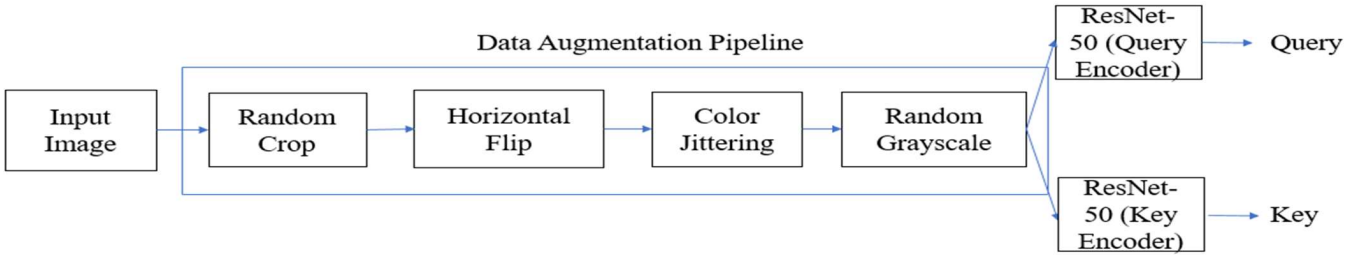
Fig. 2: MoCo Architecture

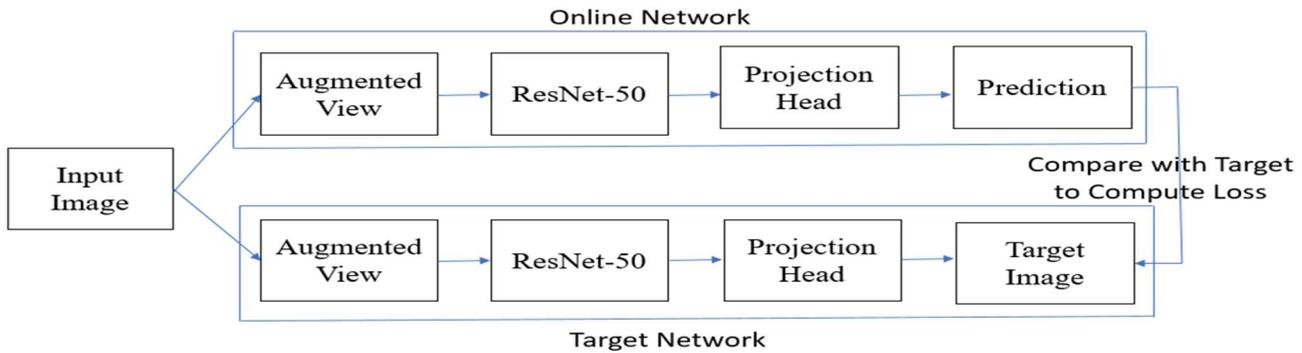
Fig. 3: BYOL Architecture

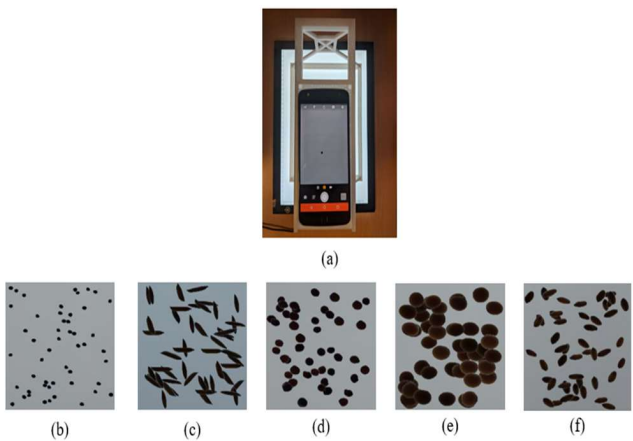
Fig. 4: (a) Image capture of soy seed on lightbox; Bottom row shows synthetic images of (b) canola (c) rough rice (d) sorghum (e) soy (f) wheat

**SimCLR:** A projection head with Linear-ReLU-Linear layers that yields 128-dimensional latent space vectors as output is added on top of the ResNet-50 model because [6] shows that the framework delivers its best performance under a non-linear projection head. A batch size of 192 is used while the network is trained. Upon training the network, the non-linear projection head is removed and each of the layers in the network is frozen. A linear classifier with one layer that predicts the class of an image is added to the network.

**MoCo:** The projection head of MoCo's is similar to that of SimCLR's and contains Linear-ReLU-Linear layers. The projection head outputs 128-dimensional latent space vectors. A dictionary of size 256 is used to keep track of the encoded keys in the 'target' network. Upon training the model, the projection head is removed and a linear classifier is added on top of the model.

**BYOL:** The projection head is made up of two linear layers where in an input vector of 2048 dimensions is taken, projected to 4096 dimensions and then, a vector reduced to 256 dimensions is output.

## V. RESULTS AND DISCUSSION

Predictions are made on SimCLR, MoCo and BYOL upon training the networks as described earlier. The accuracies obtained by the models is as shown in table I.

TABLE I

| Framework | Test Accuracy |
|---|---|
| SimCLR | 33% |
| MoCo | 39% |
| BYOL | 26% |

It is apparent from the results that the models do not generalize well to the real-world images in the test dataset. In order to enhance the performance of the models, linear classifiers are built on top of the self-supervised models and trained for 100 epochs on 5% of the labels i.e. 250 images comprising of 50 images of each seed type. 40 images of each seed type are used for training and 10 for validation. Supervised training is performed on ResNet-50 on the entire 5000 image training dataset for a total of 100 epochs. The accuracies and losses obtained by each of the models during training and validation are as shown in Fig. 5. Amongst the self-supervised learning frameworks, SimCLR and MoCo perform almost identically. It isn't too surprising considering the architectures of the models are similar in nature. SimCLR and MoCo attain a validation accuracy of ~80% and loss of ~0.6. While not highly significant in the grand scheme of things, their validation statistics are highly coherent with that of ResNet-50's. ResNet-50 also achieves a validation accuracy of ~80% and loss of ~0.6. However, the same is not true of BYOL. The model achieves a validation accuracy of ~55% and loss of ~1.1, performing well below SimCLR and MoCo. The total lack of negative examples might be the factor that contributes heavily to such a behavior. The models trained on linear classifiers are evaluated on the test dataset. The classification report containing precision, recall and F-1 score for each of the classes generated using Scikit are as shown in Fig. 6 and tabulated in table II. The terms Precision, Recall, F1-Score, Accuracy and Macro Average are briefly described below:

**Precision:** Given by $true\ positives/(true\ positives + false\ positives)$, precision indicates the number of positives correctly classified by the classifier. It is also called False Positive Rate.

**Recall:** Given by $true\ positives/(true\ positives + false\ negatives)$, recall indicates the number of positives identified by the model of all the positives in the dataset. It is also called True Positive Rate.

**F1-Score:** Considering both precision and recall, F1-Score is the harmonic mean of precision and recall. It is given by the formula, $2 * \frac{precision * recall}{precision + recall}$.

**Accuracy:** Accuracy is given by the sum of true positives and true negatives to the sum of true positives, true negatives, false positives and false negatives.

**Macro Average:** Macro Average is simply the arithmetic mean of the values for each of the classes in the dataset. It treats the contribution of each of the classes in the dataset equally.

### ResNet-50 Classification Report

|          | precision | recall | f1-score |
|----------|-----------|--------|----------|
| Canola   | 0.96      | 1.00   | 0.98     |
| Roughrice| 0.85      | 0.94   | 0.89     |
| Sorghum  | 0.79      | 0.87   | 0.83     |
| Soy      | 1.00      | 0.98   | 0.99     |
| Wheat    | 0.83      | 0.62   | 0.71     |
| accuracy |           |        | 0.90     |
| macro avg| 0.89      | 0.88   | 0.88     |

### SimCLR Classification Report

|          | precision | recall | f1-score |
|----------|-----------|--------|----------|
| Canola   | 0.60      | 1.00   | 0.75     |
| Roughrice| 0.83      | 0.39   | 0.53     |
| Sorghum  | 1.00      | 0.04   | 0.08     |
| Soy      | 0.77      | 0.82   | 0.80     |
| Wheat    | 0.32      | 0.63   | 0.43     |
| accuracy |           |        | 0.56     |
| macro avg| 0.70      | 0.58   | 0.51     |

### MoCo Classification Report

|          | precision | recall | f1-score |
|----------|-----------|--------|----------|
| Canola   | 0.83      | 1.00   | 0.91     |
| Roughrice| 0.98      | 0.75   | 0.85     |
| Sorghum  | 0.79      | 0.43   | 0.56     |
| Soy      | 0.98      | 0.91   | 0.95     |
| Wheat    | 0.43      | 0.78   | 0.55     |
| accuracy |           |        | 0.77     |
| macro avg| 0.80      | 0.78   | 0.76     |

### BYOL Classification Report

|          | precision | recall | f1-score |
|----------|-----------|--------|----------|
| Canola   | 0.65      | 0.95   | 0.77     |
| Roughrice| 0.62      | 0.46   | 0.52     |
| Sorghum  | 0.29      | 0.04   | 0.07     |
| Soy      | 0.77      | 0.89   | 0.82     |
| Wheat    | 0.43      | 0.64   | 0.51     |
| accuracy |           |        | 0.58     |
| macro avg| 0.55      | 0.60   | 0.54     |

Fig. 6: Classification Reports of ResNet-50, SimCLR, MoCo and BYOL

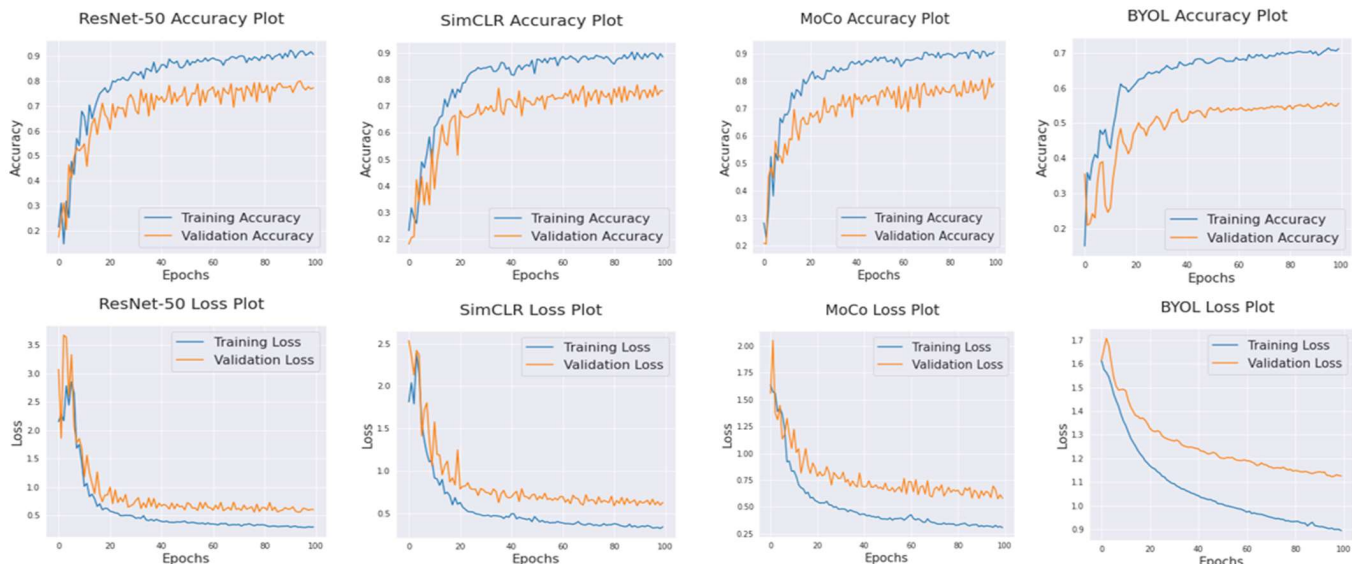

Figure 5: Training and Validation Accuracies and Losses of ResNet-50, SimCLR, MoCo and BYOL

TABLE II

|  | ResNet-50 | | | SimCLR | | | MoCo | | | BYOL | | |
|---|---|---|---|---|---|---|---|---|---|---|---|---|
|  | Precision | Recall | F1 Score | Precision | Recall | F1 Score | Precision | Recall | F1 Score | Precision | Recall | F1 Score |
| Canola | 0.96 | 1.00 | 0.98 | 0.60 | 1.00 | 0.75 | 0.83 | 1.00 | 0.91 | 0.65 | 0.95 | 0.77 |
| Rough Rice | 0.85 | 0.94 | 0.89 | 0.83 | 0.39 | 0.53 | 0.98 | 0.75 | 0.85 | 0.62 | 0.46 | 0.52 |
| Sorghum | 0.79 | 0.87 | 0.83 | 1.00 | 0.04 | 0.08 | 0.79 | 0.43 | 0.56 | 0.29 | 0.04 | 0.07 |
| Soy | 1.00 | 0.98 | 0.99 | 0.77 | 0.82 | 0.80 | 0.98 | 0.91 | 0.95 | 0.77 | 0.89 | 0.82 |
| Wheat | 0.83 | 0.62 | 0.71 | 0.32 | 0.63 | 0.43 | 0.43 | 0.78 | 0.55 | 0.43 | 0.64 | 0.51 |
|  | | | | | | | | | | | | |
| Accuracy | | | 0.90 | | | 0.56 | | | 0.77 | | | 0.58 |
| Macro Avg. | 0.89 | 0.88 | 0.88 | 0.70 | 0.58 | 0.51 | 0.80 | 0.78 | 0.76 | 0.55 | 0.60 | 0.54 |

Unsurprisingly, ResNet-50 achieves the best performance of the four models in question and establishes a steep benchmark for the self-supervised learning models to meet. Amongst the self-supervised models, MoCo is the standout winner with a macro average F1-Score that is significantly higher than SimCLR's and BYOL's. A high macro average F1-Score indicates that the model delivers good precision and recall across the five classes in the dataset. The performance of the models on the test dataset is in-line with the performance on the training dataset except for SimCLR. SimCLR's validation accuracy closely follows that of MoCo. However, the performance does not translate onto the test dataset with a macro average F1-Score of 0.51 in comparison to BYOL's 0.54, the model with the lowest validation accuracy. While achieving a macro average precision of 0.70 that is not too far from MoCo's 0.80, the macro average recall is a mere 0.58 in comparison to MoCo's 0.78 pulling down the F1-Score significantly. As for the individual seed types, the best F1-Scores are observed on canola and soy across the board with sorghum and wheat lying at the tail end. This is interesting considering the shape of the seeds. The seeds of canola, sorghum and soy are circular in nature while those of rough rice and wheat are oblong. Considering the models of SimCLR and BYOL on sorghum, a high precision and low recall is observed which indicates that the model does a good job of identifying sorghum when it does but does not think a lot of the sorghum seeds are actually sorghum. The majority of sorghum either get classified as canola or soy. While better than the performance on sorghum, a similar characteristic is observed between rough rice and wheat. Self-supervised learning frameworks have a sense of reliance on color histograms. The intuition is that the color histograms of random crops of the same image are similar and different images are dissimilar. However, random color jittering applied to images occasionally results in multiple images from different classes having similar histograms. Stark similarities between histograms of images belonging to different classes leads to error in classification. Fig. 7 shows one such case where similar histograms are obtained on color jittered images of canola and rough rice. While the images belong to two entirely different classes, the root mean square difference between the two histograms is 35.6, low enough to group both histograms in the same category. It is hard to predict the frequency of occurrence of such similarities since the data augmentation pipeline is random but it is almost certain that they will occur and hinder classification.

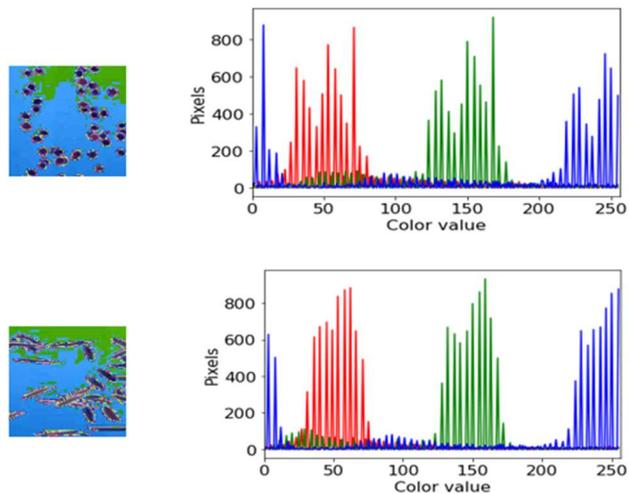

Fig. 7: Similar Color Histograms of Canola and Rough Rice

Considering the fact that every class in the dataset is a seed type and the seeds are similar to a degree, the dataset may be considered tough for self-supervised learning. In addition, the aspect of using synthetic datasets for training means that the models are supposed to learn from datasets that are not real-world. Owing to such convolutions, MoCo, the best model observed, achieves an accuracy of 77% which is only 13% shy of ResNet-50's 90% whilst being trained on a dataset that is $1/20^{th}$ the size.

## VI. CONCLUSION AND FUTURE WORK

The work evaluates the classification performance of self-supervised learning techniques namely SimCLR, MoCo and BYOL on image datasets of five different seed types namely canola, rough rice, sorghum, soy and wheat and compares it against the performance of supervised learning using ResNet-50. The results obtained demonstrate the feasibility of domain randomization and transfer learning to self-supervised models. The work presents two key ideas – a) the feasibility of synthetic datasets generated using a tiny real-world entity dataset to self-supervised learning and b) the unnecessity of a large number of labels to train linear classifiers on top of trained self-supervised models.

Moving forward, the feasibility of other types of encoders such as VGG-16 will be tested for self-supervised learning. The performance of image datasets where images are laid on different backgrounds will be tested to determine the impact of image background on the performance of the models.

## REFERENCES

[1] Rastogi, A. Reducing computational constraints in SimCLR using Momentum Contrast V2 (MoCo-V2) in PyTorch [Blog Post]. Retrieved from: https://medium.com/analytics-vidhya/simclr-with-less-computational-constraints-moco-v2-in-pytorch-3d8f3a8f8bf2

[2] Rastogi, A. Understanding SimCLR – A Simple Framework for Contrastive Learning of Visual Representations with Code [Blog Post]. Retrieved from: https://medium.com/analytics-vidhya/understanding-simclr-a-simple-framework-for-contrastive-learning-of-visual-representations-d544a9003f3c

[3] Buckland, M., & Gey, F. (1994). The relationship between recall and precision. *Journal of the American society for information science*, *45*(1), 12-19.

[4] Bouchard, Louis. What is Self-Supervised Learning? Will machines ever be able to learn like humans? [Blog Post]. Retrieved from: https://medium.com/what-is-artificial-intelligence/what-is-self-supervised-learning-will-machines-be-able-to-learn-like-humans-d9160f40cdd1

[5] Chen, X., Fan, H., Girshick, R., & He, K. (2020). Improved baselines with momentum contrastive learning. *arXiv preprint arXiv:2003.04297*.

[6] Chen, T., Kornblith, S., Norouzi, M., & Hinton, G. (2020, November). A simple framework for contrastive learning of visual representations. In *International conference on machine learning* (pp. 1597-1607). PMLR.

[7] Ghosal, S., Zheng, B., Chapman, S. C., Potgieter, A. B., Jordan, D. R., Wang, X., & Ganapathysubramanian, B. (2019). A weakly supervised deep learning framework for sorghum head detection and counting. Plant Phenomics, 2019, 1525874.

[8] Grill, J. B., Strub, F., Altché, F., Tallec, C., Richemond, P. H., Buchatskaya, E., ... & Valko, M. (2020). Bootstrap your own latent: A new approach to self-supervised learning. *arXiv preprint arXiv:2006.07733*.

[9] Gulzar, Y., Hamid, Y., Soomro, A. B., Alwan, A. A., & Journaux, L. (2020). A Convolution Neural Network-Based Seed Classification System. *Symmetry*, *12*(12), 2018.

[10] He, K., Fan, H., Wu, Y., Xie, S., & Girshick, R. (2020). Momentum contrast for unsupervised visual representation learning. In *Proceedings of the IEEE/CVF Conference on Computer Vision and Pattern Recognition* (pp. 9729-9738).

[11] Hendrycks, D., Mazeika, M., Kadavath, S., & Song, D. (2019). Using self-supervised learning can improve model robustness and uncertainty. *arXiv preprint arXiv:1906.12340*.

[12] Kamilaris, A., & Prenafeta-Boldú, F. X. (2018). Deep learning in agriculture: A survey. Computers and electronics in agriculture, 147, 70-90.

[13] Kaiming He, Xiangyu Zhang, Shaoqing Ren, and Jian Sun. Deep residual learning for image recognition. In CVPR, 2016.

[14] Kaiming He, Xiangyu Zhang, Shaoqing Ren, and Jian Sun. Identity mappings in deep residual networks. In ECCV, 2016.

[15] Li, L., Zhang, Q., & Huang, D. (2014). A review of imaging techniques for plant phenotyping. Sensors, 14(11), 20078-20111.

[16] Lin, Y., Tang, C., Chu, F. J., & Vela, P. A. (2020, May). Using Synthetic Data and Deep Networks to Recognize Primitive Shapes for Object Grasping. In 2020 IEEE International Conference on Robotics and Automation (ICRA) (pp. 10494-10501). IEEE.

[17] Mahmood, A., Bennamoun, M., An, S., Sohel, F., Boussaid, F., Hovey, R., & Kendrick, G. (2020). Automatic detection of Western rock lobster using synthetic data. ICES Journal of Marine Science, 77(4), 1308- 1317.

[18] Margapuri, V., & Neilsen, M. (2020). Seed Phenotyping on Neural Networks using Domain Randomization and Transfer Learning. *arXiv preprint arXiv:2012.13259*.

[19] Misra, I., & Maaten, L. V. D. (2020). Self-supervised learning of pretext-invariant representations. In *Proceedings of the IEEE/CVF Conference on Computer Vision and Pattern Recognition* (pp. 6707-6717).

[20] Mohanty, S. P., Hughes, D. P., & Salathé, M. (2016). Using deep learning for image-based plant disease detection. Frontiers in plant science, 7, 1419.

[21] Nafi, N. M., & Hsu, W. H. (2020, July). Addressing Class Imbalance in Image-Based Plant Disease Detection: Deep Generative vs. Sampling-Based Approaches. In *2020 International Conference on Systems, Signals and Image Processing (IWSSIP)* (pp. 243-248). IEEE.

[22] Nai, A. SimCLR: Contrastive Learning of Visual Representations [Blog Post]. Retrieved from: https://medium.com/@nainaakash012/simclr-contrastive-learning-of-visual-representations-52ecf1ac11fa

[23] Odom, F. Easy Self-Supervised Learning with BYOL [Blog Post]. Retrieved from: https://medium.com/the-dl/easy-self-supervised-learning-with-byol-53b8ad8185d

[24] Oord, A. V. D., Li, Y., & Vinyals, O. (2018). Representation learning with contrastive predictive coding. *arXiv preprint arXiv:1807.03748*.

[25] Peng, X. B., Andrychowicz, M., Zaremba, W., & Abbeel, P. (2018, May). Sim-to-real transfer of robotic control with dynamics randomization. In 2018 IEEE international conference on robotics and automation (ICRA) (pp. 1-8). IEEE.

[26] Priya Goyal, Dhruv Mahajan, Abhinav Gupta, and Ishan Misra. Scaling and benchmarking self-supervised visual representation learning. In ICCV, 2019.

[27] Sur, R. SimCLR – A Simple Framework for Constrastive Learning of Visual Representation [Blog Post]. Retrieved from: https://rittikasur16.medium.com/simclr-a-simple-framework-for-contrastive-learning-of-visual-representation-799a903c2779

[28] Sajjadi, M. S., Bachem, O., Lucic, M., Bousquet, O., & Gelly, S. (2018). Assessing generative models via precision and recall. *arXiv preprint arXiv:1806.00035*.

[29] Santos, L., Santos, F. N., Oliveira, P. M., & Shinde, P. (2019, November). Deep learning applications in agriculture: A short review. In *Iberian Robotics conference* (pp. 139-151). Springer, Cham.

[30] Tobin, J., Fong, R., Ray, A., Schneider, J., Zaremba, W., & Abbeel, P. (2017, September). Domain randomization for transferring deep neural networks from simulation to the real world. In *2017 IEEE/RSJ international conference on intelligent robots and systems (IROS)* (pp. 23-30). IEEE.

[31] Toda, Y., Okura, F., Ito, J., Okada, S., Kinoshita, T., Tsuji, H., & Saisho, D. (2020). Training instance segmentation neural network with synthetic datasets for crop seed phenotyping. *Communications biology*, *3*(1), 1-12.

[32] Toda, Y., Okura, F., Ito, J., Okada, S., Kinoshita, T., Tsuji, H., & Saisho, D. (2019). Learning from synthetic dataset for crop seed instance segmentation. *BioRxiv*, 866921.

[33] Toth, D., Cimen, S., Ceccaldi, P., Kurzendorfer, T., Rhode, K., & Mountney, P. (2019, May). Training deep networks on domain randomized synthetic X-ray data for cardiac interventions. In International Conference on Medical Imaging with Deep Learning (pp. 468-482). PMLR.

[34] USDA. Ag and Food Sectors and the Economy [Blog Post]. Retrieved from: https://www.ers.usda.gov/data-products/ag-and-food-statistics-charting-the-essentials/ag-and-food-sectors-and-the-economy/

[35] Wu, L., Hoi, S. C., & Yu, N. (2010). Semantics-preserving bag-of-words models and applications. *IEEE Transactions on Image Processing*, *19*(7), 1908-1920.

[36] Yann LeCun. Predictive learning, 2016. URL https://www.youtube.com/watch?v= Ount2Y4qxQo.

[37] Zakharov, S., Kehl, W., & Ilic, S. (2019). Deceptionnet: Network-driven domain randomization. In *Proceedings of the IEEE/CVF International Conference on Computer Vision* (pp. 532-541).